\documentclass[10pt,twocolumn,letterpaper]{article}

\usepackage[pagenumbers]{cvpr} 

\usepackage[dvipsnames]{xcolor}

\definecolor{cvprblue}{rgb}{0.21,0.49,0.74}
\usepackage[pagebackref,breaklinks,colorlinks,citecolor=cvprblue]{hyperref}

\usepackage{multirow}
\usepackage{bbding}

\title{MegActor: Harness the Power of Raw Video for Vivid Portrait Animation}

\author {
    Shurong Yang\footnotemark[1]\ ,
    Huadong Li\footnotemark[1]\ ,
    Juhao Wu\footnotemark[1]\ ,
    Minhao Jing\footnotemark[1] \footnotemark[2],\\
    Linze Li, 
    Renhe Ji\footnotemark[3],
    Jiajun Liang\footnotemark[3],
    Haoqiang Fan, \\
    MEGVII Technology \\
    \tt\small $\{$yangshurong6894, jingminhao666$\}$@gmail.com \\ 
    \tt\small $\{$lihuadong, wujuhao, lilinze, jirenhe, liangjiajun, fanhaoqiang$\}$@megvii.com \\
    \href{https://github.com/megvii-research/MegActor}{megvii-research/MegActor} 
}

\begin{document}

\twocolumn[{%
\maketitle
\begin{center}
    \centering
    \includegraphics[width=0.82\linewidth]{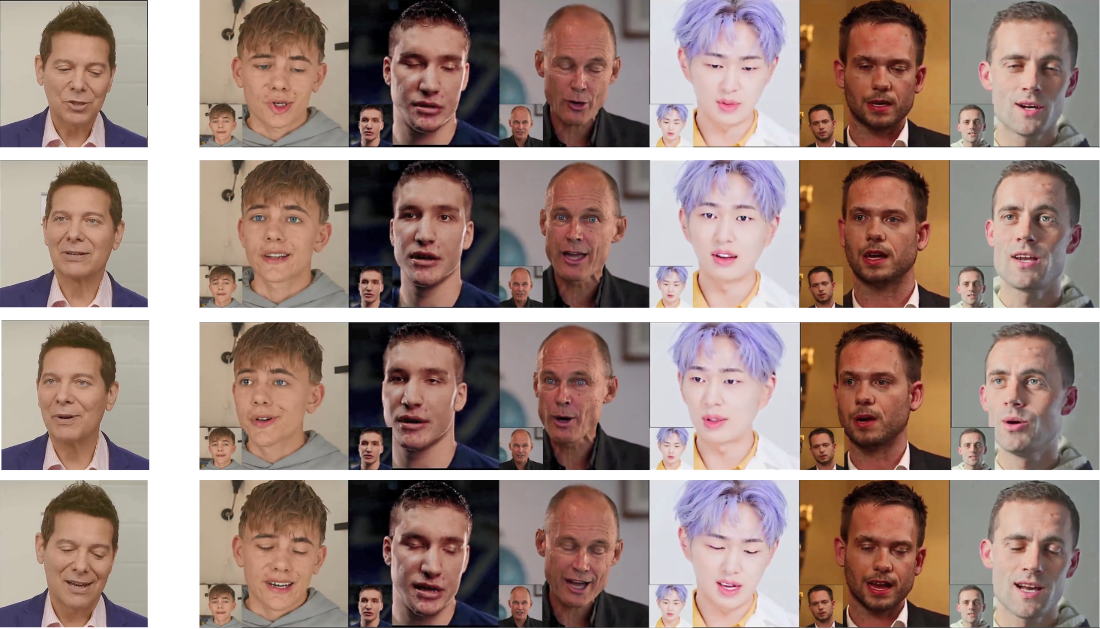}
    \captionof{figure}[htbp]{
    Given a raw driving video (left column), MegActor can synthesize captivating and expressive animations (right column), encompassing the head pose variations and detailed facial expressions present in the input driving video with multiple reference portraits.
    }
\label{fig:visual}
\end{center}%
}]
\maketitle


\begin{abstract}
Despite raw driving videos contain richer information on facial expressions than intermediate representations such as landmarks in the field of portrait animation, they are seldom the subject of research.
This is due to two challenges inherent in portrait animation driven with raw videos: 1) Significant identity leakage; 
2) Irrelevant background and facial details such as wrinkles degrade performance.
To harnesses the power of the raw videos for vivid portrait animation, we proposed a pioneering conditional diffusion model named as MegActor.
First, we introduced a synthetic data generation framework for creating videos with consistent motion and expressions but inconsistent IDs to mitigate the issue of ID leakage. 
Second, we segmented the foreground and background of the reference image and employed CLIP to encode the background details. This encoded information is then integrated into the network via a text embedding module, thereby ensuring the stability of the background.
Finally, we further style transfer the appearance of the reference image to the driving video to eliminate the influence of facial details in the driving videos.
Our final model was trained solely on public datasets, achieving results comparable to commercial models. We hope this will help the open-source community.
The code is available at \href{https://github.com/megvii-research/MegActor}{\textit{https://github.com/megvii-research/MegActor}}.

\end{abstract}

{
  \renewcommand{\thefootnote}{\fnsymbol{footnote}}
  \footnotetext[1]{Indicates equal contribution.}
  \footnotetext[2]{Lead this project.}
  \footnotetext[3]{Corresponding author.}
}

\section{Introduction}
\label{sec:Intro}


Portrait Animation is a task to transfer the motion and facial expressions from a driving video to a target portrait while preserving the identity and background of the target portrait, which has many potential applications such as digital avatars~\cite{wang2021one,ma2021pixel}, AI-based human conversations~\cite{aicommunicate,johnson2018assessing,kessler2018technology}, etc.
Beginning with the advent of GANs~\cite{goodfellow2014generative} and NeRF~\cite{mildenhall2021nerf}, numerous studies have delved into the fields of portrait animation~\cite{bansal2018recycle,ha2020marionette,kim2018deep,mildenhall2021nerf,hong2022headnerf,gao2022reconstructing,chan2022efficient,chen2022tensorf}.
However, these generated methods often produce unrealistic and distorted faces, accompanied by artifacts such as blurring and flickering.

\begin{table*}[]
    \centering
    \resizebox{0.99\textwidth}{!}{
    \begin{tabular}{l c c c c | c c}
    \toprule
         Model & Backbone & Control Signal & Public Dataset & Dataset Size  & Open Code & Open Weights \\
         \midrule
         Animate Anyone\cite{hu2023animate} & SD1.5 & Sketch & \textcolor{red}{$\times$} & 10 Hours & \textcolor{red}{$\times$} & \textcolor{red}{$\times$} \\
         DisCo\cite{wang2023disco} & SD1.5 & Sketch & \textcolor{green}{$\checkmark$} & 1.5 Hours & \textcolor{green}{$\checkmark$} & \textcolor{green}{$\checkmark$} \\
         MagicAnimate\cite{xu2023magicanimate} & SD1.5 & Sketch & \textcolor{green}{$\checkmark$} & 10 Hours & \textcolor{green}{$\checkmark$} & \textcolor{green}{$\checkmark$} \\
        MagicPose\cite{chang2023magicdance} & SD1.5 & Sketch + Landmark & \textcolor{green}{$\checkmark$} & 1.5 Hours & \textcolor{green}{$\checkmark$} & \textcolor{green}{$\checkmark$} \\
              \midrule
         EMO \cite{tian2024emo} & SD1.5 & Audio + Weak Condition & \textcolor{red}{$\times$} & 250 Hours & \textcolor{red}{$\times$} & \textcolor{red}{$\times$} \\   
         VASA\cite{xu2024vasa} & Motion Diffusion & Audio + Weak Condition & \textcolor{red}{$\times$} & Unknown & \textcolor{red}{$\times$} & \textcolor{red}{$\times$} \\
    \midrule
        X-portrait\cite{xie2024x} & SD1.5 & Raw Video & \textcolor{red}{$\times$} & 180 Hours & \textcolor{red}{$\times$} & \textcolor{red}{$\times$} \\
        MegActor (Ours) & SD1.5 & Raw Video & \textcolor{green}{$\checkmark$} & 700 Hours & \textcolor{green}{$\checkmark$} & \textcolor{green}{$\checkmark$} \\
    \bottomrule
    \end{tabular}}
    \caption{Comparison with SoTA methods. MegActor is a method for vivid portrait animation with raw driving video, utilizing only public datasets and releasing the code and model weights.
    }
    \label{tab:compare}
\end{table*}

In recent years, Stable Diffusion(SD) models~\cite{rombach2022high} have demonstrated their advantages in creating high-quality images and videos.
Researchers have attempted to utilize stable diffusion models in portrait animation tasks. 
These methods can be divided into three major categories based on control signals: text to video (T2V), image to video (I2V), and audio to video (A2V).
T2V methods~\cite{han2023generalist,liu2024towards,xu2024facechain} encode identity, motion information, and background content from the reference image and drive frame with CLIP~\cite{radford2021learning} and ArcFace~\cite{deng2019arcface}, replace the text embeddings in the stable diffusion model and inject them into the network via cross-attention.
This coarse control mechanism struggles to manage subtle movements, such as those of the eyebrows, lips, and eyelids, and fails to reconstruct pixel-level details in the background.

I2V methods~\cite{hu2023animate,xu2023magicanimate,chang2023magicdance,zhu2024champ,wang2023disco} leverages general control signals, such as facial landmarks, dense poses~\cite{guler2018densepose}, and skeletons, to accurately extract motion information from the driving video and transform it into the reference image for reconstruction.
However, these methods are constrained by the limitations of generic control signals, which fail to animate subtle facial expressions, such as forehead and eye movements. 
Furthermore, the generation of general control signals heavily relies on the robustness and accuracy of third-party pose detectors, such as DWPose~\cite{yang2023effective} and DensePose~\cite{guler2018densepose}, hindering the animation's expressiveness and stability. 
A2V methods~\cite{tian2024emo,xu2024vasa,wei2024aniportrait} employs audio control signals to drive lip movement in the reference image, supplemented by weak control signals for head motion. These approaches mitigates facial distortion issues and produces natural portrait animation videos. However, it still cannot control subtle facial movements, such as eye movements, and often results in head movements that are inconsistent with the amplitude of the driving video.

Intuitively, driven by raw video can generate more subtle motion and expression animations, such as eyeball and head movement, compared to the control signals of the previous intermediate representation.
Despite the raw driving video containing richer facial expression information, it still presents two challenges: 1) significant identity leakage; 
2) Irrelevant background and facial details such as wrinkles degrade performance.

To harnesses the power of the raw videos for vivid portrait animation, we proposed MegActor, a pioneering conditional diffusion model that addresses the above challenges, showcasing superior generation quality, consistent animation, identity  preservation, and strong generalization capabilities, as shown in Fig. \textcolor{red}{1}.
When using raw video as the driving source, the model is highly prone to directly replicating the appearance from the raw video, leading to identity leakage, as the reference images are extracted from frames of the driving video during training.
Therefore, we designed a synthetic data generation framework to create videos with consistent motion and expressions but inconsistency identity. This framework includes face-swapping, stylization \cite{podell2023sdxl} and randomly transforming the raw driving video.
Subsequently, the raw driving video introduces unexpected noise from background and facial details such as wrinkles. To address this, we segment the  background of the reference image and use CLIP to encode its background details. This encoded information is then integrated into the network via a text embedding module, ensuring the stability of the background. we further style transfer the appearance of the reference image to the driving video to eliminate
the influence of facial details in the driving videos.

Our model is trained exclusively on public datasets to ensure reproducibility of the results. A comprehensive comparison table with SOTA methods can be found in Tab. \ref{tab:compare}.
As shown in Fig. \ref{fig:visual}, the visualization results demonstrate that our approach can synthesize natural and expressive portrait animations, encompassing the head pose variations and detailed facial expressions present in the raw driving video.
Our contributions are summarized as follows:
\begin{itemize}
\item  A novel portrait animation approach with raw driving video control, employing a synthetic data generation framework, effectively maximizes motion consistency and mitigates identity leakage issues.
\item Enhancing robustness to irrelevant information in raw driving video, with the stylization of reference image to raw driving frames.
\item The experimental results show that our approach, trained on public datasets, achieves comparable outcomes to commercial models, demonstrating the effectiveness of our approach.
\end{itemize}

\section{Related Work}
\label{sec:Related}

\subsection{GAN-based Portrait Animation}

A majority of portrait animation methods utilize generative adversarial networks (GANs) to learn motion dynamics in a self-supervised manner.
Researches has primarily focused on leveraging latent encodings\cite{drobyshev2022megaportraits,hong2022depth,hong2023implicit,ma2024cvthead,zhang2023metaportrait}, 2D keypoints\cite{liu2021self,siarohin2019first,wang2021one,zakharov2019few,zhao2022thin,zhao2021sparse}, and 3D facial prior models\cite{tran2018nonlinear,wu2021f3a,yao2020mesh,zhu2017face} (e.g., 3DMM\cite{blanz2023morphable}), to separate motion features from identity features. Techniques such as GANs\cite{bansal2018recycle,ha2020marionette,kim2018deep,wang2018high,yang2020transmomo,zakharov2019few}, Nerf\cite{mildenhall2021nerf,hong2022headnerf,gao2022reconstructing,chan2022efficient,chen2022tensorf,liu2020neural,muller2022instant,fridovich2022plenoxels,sun2022direct}, and motion decoders\cite{park2019semantic,wang2018high} have been used as renders to generate high-fidelity facial animations. However, due to limitations in the capabilities of motion transformation networks (typically composed of simple 2D convolutions and MLPs) and renders, these methods often produce unrealistic and distorted faces, accompanied by artifacts such as blurring and flickering.

\subsection{Diffusion-based Portrait Animation}

Stable Diffusion (SD) models have shown their superior performance in high-quality image and video creation. Researchers are exploring stable diffusion in portrait animation, categorized into T2V, I2V, and A2V based on control signals.
T2V methods~\cite{han2023generalist,liu2024towards,xu2024facechain} encode identity and motion from reference images and driving frames using CLIP~\cite{radford2021learning} and ArcFace~\cite{deng2019arcface}, integrating them into the SD model via cross-attention. However, they struggle with subtle movements and background details.
I2V methods~\cite{hu2023animate,xu2023magicanimate,chang2023magicdance,zhu2024champ,wang2023disco} use facial landmarks and poses to extract motion for reconstruction but are limited in animating delicate expressions and rely on external pose detectors for control signal generation, affecting expressiveness and stability.
A2V methods~\cite{tian2024emo,xu2024vasa,wei2024aniportrait} use audio signals for lip synchronization and weak signals for head motion, reducing facial distortion and producing natural videos, yet they can't control nuanced eye movements and may mismatch head motion amplitude to the driving video.

\section{Dataset Pipline}
We utilized only publicly available datasets, VFHQ\cite{xie2022vfhq} and CelebV-HQ\cite{zhu2022celebv}, for training. During training, we randomly selected a frame as the first frame and sampled multiple frames with a certain stride to form a segment of video used for training. For each video segment, we randomly selected one frame as the reference image, and all frames were used as the driving video and the ground truth for the model to fit.

However, directly using all frames as the driving video may lead to leakage of identity information and irrelevant background information. This is because the public datasets lack videos where different individuals perform the same actions, and the original videos serve as both the driving video and the ground truth in training, resulting in the model simply replicating the driving video as the generated result.

To prevent identity leakage during training, manifested by the model producing results identical to the driving video due to identical identities between the driving video and the ground truth, we generated a portion of AI face-swapping data using Face-Fusion from ModelScope \ref{Face-FusionSection} and synthesized a portion of stylized data using SDXL\cite{podell2023sdxl} \ref{StylizingSection}. To better control eye movements, we utilized L2CSNet\cite{abdelrahman2023l2cs} to select a portion of data with significant eye movement amplitudes for fine-tuning the model \ref{DataFilteringSection}. We augmented the driving videos to address issues of identity leakage and background leakage, as well as to enhance the model's generalization to different facial shapes \ref{DataAugmentationSection}.

\subsection{Face-Fusion}\label{Face-FusionSection}
We took each frame of the driving video as the target image and paired it with another image from different videos in the dataset featuring a different individual as the source image. Our face-swapping model then transferred the identity from the source image onto the target image. The resulting generated images were utilized as frames in the driving video. The model was fitted to the original real video based on the driving video and the reference image, rather than the face-swapped driving video. This ensured that the model's prediction target was always to generate real faces, thereby enhancing the realism of our generated results.

\subsection{Stylizing}\label{StylizingSection}
We applied SDXL\cite{podell2023sdxl} to individually stylize each frame of a portion of the videos, and the resulting stylized video was used as the driving video during training. Although the stylized video underwent significant changes in both the identity information and background compared to the original video, the actions of the individuals remained almost unchanged. Although the continuity between frames of the generated results was disrupted, the model was fitted to the original real video during training. Therefore, using stylized videos as the driving video during training did not affect the continuity of the final generated results.

\subsection{Data Filtering}\label{DataFilteringSection}
We employed L2CSNet\cite{abdelrahman2023l2cs} to assess changes in gaze between frames and filtered out videos with significant gaze changes. Approximately 5$\%$ of the total data were identified as having significant gaze changes. After training the model in the first stage using all of data, we conducted fine-tuning using the data with significant gaze changes.

\subsection{Data Augmentation}\label{DataAugmentationSection}
We utilized pyFacer\cite{deng2020retinaface} to detect faces in each frame of the videos and retained only the facial regions from the driving videos, setting all pixels outside the facial region to black to reduce leakage of background information during training. To prevent identity leakage resulting from identical driving and real videos, we applied random augmentations to the driving videos. Augmentation techniques included grayscale conversion and random adjustments to size and aspect ratio, affecting only facial shape without altering facial expressions or head poses. Modifying facial shapes also provided additional training samples where the driving videos differed from the reference image in terms of facial structure, enhancing the model's generalization to different facial shapes. The model received both the reference image and the augmented driving videos and was trained to fit the real videos.

\section{Method}

\begin{figure*}[htbp]
    \centering
    \includegraphics[width=0.99\textwidth]{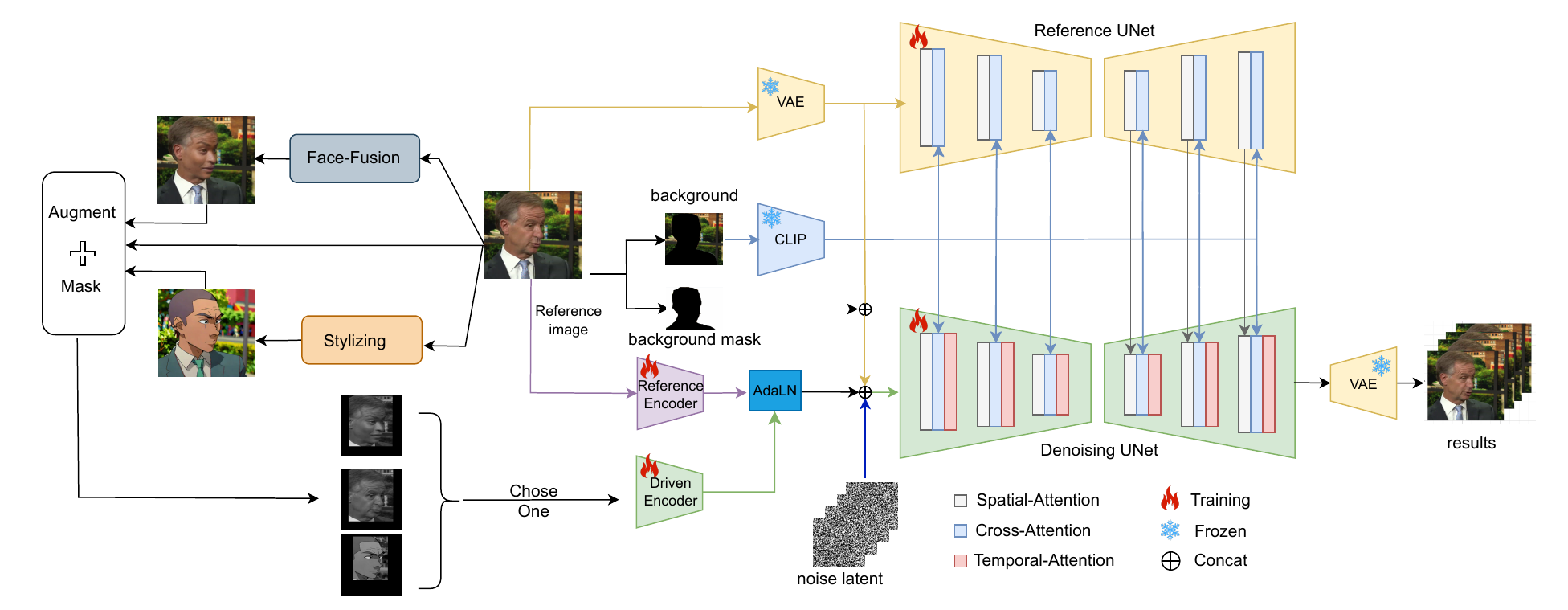}
    \caption{Overview of the proposed method. The raw video frames are processed by AI face-swapping and stylization to modify the character ID, then data augmentation methods such as scaling and aspect ratio adjustment are applied, and finally, all parts except the face are masked out to obtain the driving video, which is then fed into the DrivenEncoder. The encoding results of the DrivenEncoder are concatenated along the channel dimension with latent noise, the latent code of the reference image, and the foreground and background masks, and then fed into the Denoising UNet. MegActor's ReferenceNet extracts identity and background information of the character and injects this information into the Denoising UNet through cross-attention. CLIP encodes the reference image background and replaces the text embedding to be injected into the ReferenceNet and Denoising UNet.}
    \label{fig:method}
\end{figure*}

Given a reference image containing a static face, our goal is to use a driving face video to animate the reference image, making the character in the reference image perform the same actions as those in the driving video. The generated result should retain the identity of the character from the reference image and preserve the background from the reference image.

In this work, we use SD1.5~\cite{rombach2022high} as the pre-trained denoising model. We use a UNet network with the same architecture as SD\cite{rombach2022high}, referred to as ReferenceNet, to extract fine-grained identity and background information from the reference image \ref{ReferenceNetSection}. For the driving video, we employ a DrivenEncoder containing multiple layers of 2D convolutions to extract motion features \ref{DrivenEncoderSection}. 
To enhance the continuity between generated frames, we insert a temporal module into the denoising model and fine-tune the temporal module separately \ref{TemporalLayerSection}. We encode the background from the reference image using CLIP~\cite{radford2021learning}'s image encoder and inject this information into the model \ref{ImageEncoderSection}.We provide several training techniques to mitigate discrepancies between training and inference phases.
Fig. \ref{fig:method} provides an overview for the model architecture with the training pipeline.

\subsection{ReferenceNet}\label{ReferenceNetSection}
In recent works\cite{hu2023animate,xu2023magicanimate,chang2023magicdance,zhu2024champ,wang2023disco}, it has been discovered that the intermediate features of diffusion models possess remarkable communication capabilities, enabling pixel-level fine control of images. We employ a UNet with an identical architecture to the Denoising UNet network, referred to as ReferenceNet, to extract features from the reference image. The reference image is first encoded by the VAE\cite{kingma2013auto}, transforming it into latent feature representations,  $R_l$
. Subsequently, ReferenceNet extracts spatial features from the $R_l$ and injects the intermediate features into the Denoising UNet.

Unlike \cite{hu2023animate,chang2023magicdance}, which inject spatial features from all layers into the Denoising UNet, we only inject features from the mid-layer and up-block layers into the Denoising UNet.  This design allows the down-block of the Denoising UNet to fully integrate motion features extracted by the DrivenEncoder.

\subsection{DrivenEncoder}\label{DrivenEncoderSection}
Inspired by \cite{hu2023animate}, we employ a lightweight DrivenEncoder, which utilizes four 2D convolutional layers (4×4 kernels, 2×2 strides, with 16, 32, 64, and 128 channels) to extract motion features from the driving video. These motion features are then aligned to the same resolution as the noise latents obtained from random sampling. We do not use ControlNet\cite{zhang2023adding}, which reduces the computational load.

We concatenate the motion features with the noise latents along the channel dimension, and during training, we reinitialize the parameters of the conv-in layer of the Denoising UNet. We retain the parameters of the first four channels and initialize the parameters of the remaining channels to zero. This approach mitigates the disruption to the spatial structure of the Denoising UNet originally initialized from SD1.5 due to the introduction of motion features.

Many studies have shown that different individuals have unique speaking styles\cite{tan2024say,tan2024style2talker,guan2023stylesync,tan2024edtalk}. Injecting only the motion features from the driving video into the network can lead to over-rigid facial movements due to direct motion transfer, thereby reducing the realism of the generated results. 
Therefore, we again use the reference image as a reference for the DrivenEncoder and the down-block of the Denoising UNet when extracting motion features from the driving video.
We input the latent representation obtained from encoding the reference image with the VAE\cite{kingma2013auto}, along with a mask used to segment the foreground character, extracted using DensePose\cite{guler2018densepose}, into the Denoising UNet.The latent representation of the reference image and the foreground mask are concatenated with the noise latents and the motion features extracted by the DrivenEncoder along the channel dimension, and then fed into the Denoising UNet.The parameters of the newly added channels in the conv-in layer of the Denoising UNet are initialized to zero.

We also incorporate the speaking style of the character into the motion features extracted by the DrivenEncoder.Inspired by AdaLN\cite{huang2017arbitrary}, we use a 2D convolution with the same architecture as the DrivenEncoder to encode the reference image and use an Multi-layer perceptron to transform the reference image features into $scale$ and $shift$ parameters, which are applied to the motion features along the channel dimension.

\subsection{Temporal Layer}\label{TemporalLayerSection}
AnimateDiff\cite{guo2023animatediff} demonstrates that inserting additional time modules into Text-to-Image (T2I) models in video generation tasks can capture temporal dependencies between video frames and enhance the continuity between them. This design facilitates the transfer of pre-trained image generation capabilities from the base T2I model. Inspired by AnimateDiff, we insert a time module after each Res-Trans layer of the Denoising UNet to perform temporal attention between frames.

\subsection{ImageEncoder}\label{ImageEncoderSection}
We utilize the image encoder from CLIP\cite{radford2021learning} as an alternative to the text encoder in cross-attention. The image encoder transforms input images into multiple patches, treating each patch as tokens similar to those in text for feature extraction. Additionally, a special token, $CLS$, is used to represent global features. We employ the image encoder to encode the background portion of the reference image. The extracted global features represented by $CLS$ and the local features represented by each patch are merged to replace text embedding, which will be fed into the Denoising UNet and ReferenceNet using cross-attention mechanisms.

\begin{figure*}[htbp]
    \centering
    \includegraphics[width=0.99\textwidth]{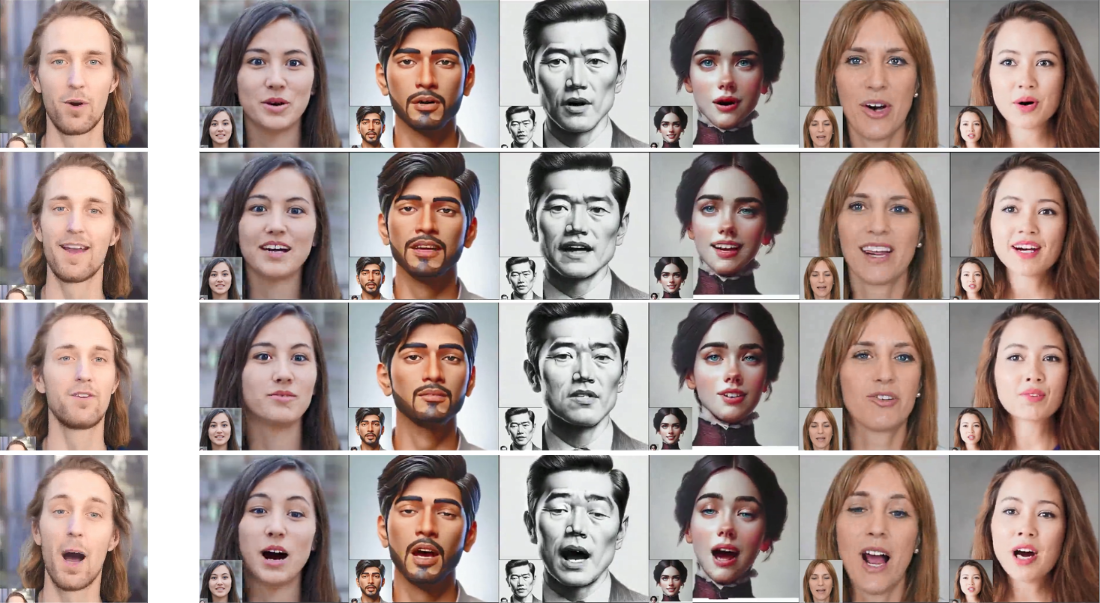}
    \caption{Visualization results. To further demonstrate the generalizability of our approach, we used the official generated results from VASA~\cite{xu2024vasa} as the driving video (left column) to animate multiple reference images  (right column) from VASA~\cite{xu2024vasa}. The enriched and realistic generated videos, encompassing consistent expressions and head movements, showcase the robustness of our approach.
    }
    \label{fig:results-self}
\end{figure*}




\section{Experiments}
\label{experiment}

\subsection{Implementation Details}\label{TrainingDetailSection}

\noindent\textbf{Dataset.}
We utilized publicly available datasets for training, including VFHQ~\cite{xie2022vfhq} and
CelebV-HQ~\cite{zhu2022celebv}. During training.
To prevent identity leakage during training, we generated a portion of AI face-swapping data using Face-Fusion from ModelScope and synthe-
sized a portion of stylized data using SDXL~\cite{podell2023sdxl}. 
To better control eye movements, we utilized L2CSNet~\cite{abdelrahman2023l2cs} to select data with significant eye movement amplitudes for fine-tuning the model.
For the benchmark, we utilized the official test cases from VASA~\cite{xu2024vasa} and EMO~\cite{tian2024emo}, along with additional out-of-domain portrait images that we collected.

\noindent\textbf{Training Details.}
Our model training consists of two stages. In the first stage, the model does not include a Temporal Layer. We freeze the ImageEncoder and only train the Driven Encoder, Denoising UNet, and ReferenceNet. In the second stage, we insert the Temporal Layer into the Denoising UNet. The parameters of the Temporal Layer are initialized from AnimateDiff, and only the Temporal Layer is trained.
During both the first and second stage training, we use a mix of AI face-swapping data, stylized data, and real data as driving videos, with respective proportions of 40$\%$, 10$\%$, and 50$\%$. The data sampling stride is set to 2, and the length of the training videos is set to 16 frames. After the first stage training, we fine-tune the model using videos filtered to have significant gaze changes. The data sampling stride is set to 12, and the length of the training videos remains at 16 frames.
We conduct training on 8 V100 GPUs. All video frames are resized to 512x512. Throughout the entire training process, we use the AdamW optimizer with a constant learning rate of 1e-5.

\noindent\textbf{Inference Details.} 
During the inference phase, we implement an overlapping sliding window approach to generate long videos. We infer 16 frames of video at a time with an overlap of 8 frames, taking the average of the two generations for the overlapping regions as our final result.

\begin{figure*}[htbp]
    \centering
    \includegraphics[width=0.99\textwidth]{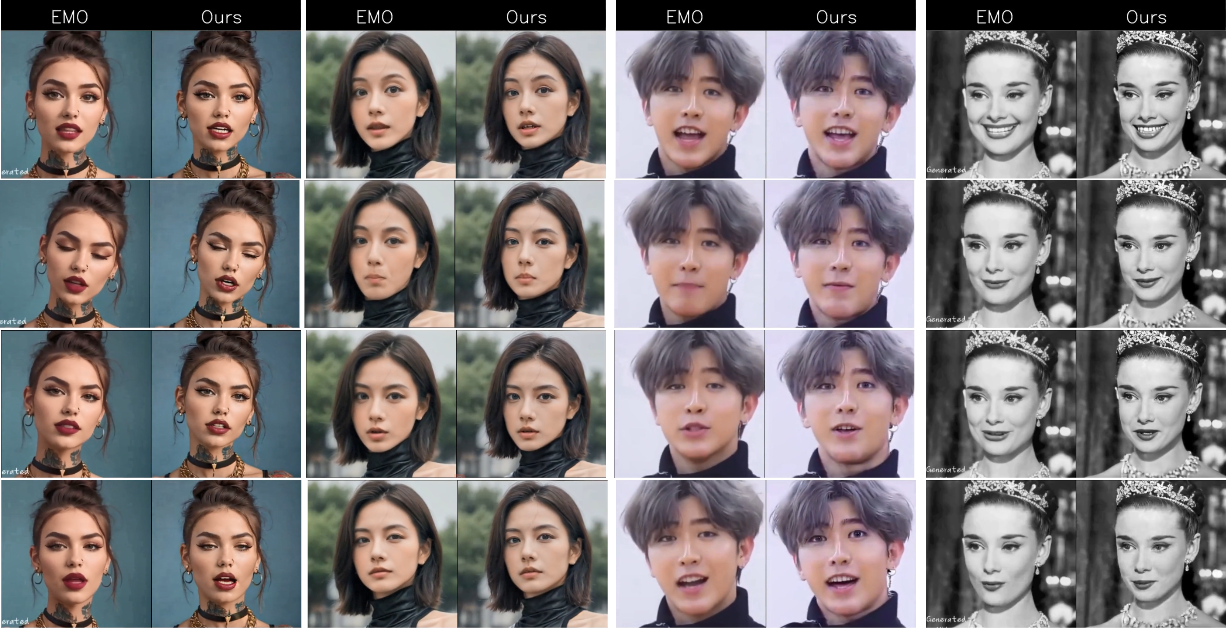}
    \caption{Compared with SOTA portrait animation method EMO~\cite{tian2024emo}. Since EMO has not released its inference code, we selected cases from EMO's official demonstration for comparison. The visualization results show that our method achieves comparable effects to EMO.}
    \label{fig:results-emo}
\end{figure*}

\subsection{Evaluations and Comparisons}
We tested our model on Cross-identity data. We extracted a segment of video from VFHQ~\cite{xie2022vfhq} as the driving video and selected six other frames  from different identities as reference images. The generated results are shown in the figure. The backgrounds in the results match the reference images at the pixel level, and the six individuals maintain the same identity information as the reference images while performing the same facial expressions and head movements as those in the driving video, including subtle movements such as eye motion. This demonstrates MegActor's excellent capability in generating portrait animations under cross-ID conditions.

We also extracted a segment of video from the VASA~\cite{xu2024vasa} test samples as the driving video and selected the first frame from six other VASA~\cite{xu2024vasa} test samples as reference images. The results show that MegActor can produce realistic outputs even in Cross-identity tests on the VASA~\cite{xu2024vasa} test samples.

In the figure, we compare our results with the state-of-the-art EMO~\cite{tian2024emo}. We used the test sample videos from EMO~\cite{tian2024emo} as the driving videos and the first frame of these videos as the reference images. The results show that some frames generated by EMO~\cite{tian2024emo} exhibit blurriness in areas such as teeth, whereas MegActor's results have clearer teeth. This comparison indicates that MegActor can achieve comparable results to EMO~\cite{tian2024emo}.


\subsection{Limitations and Future Work}
We aim to refine the capability of MegActor to generate consistent videos in the future, especially in intricate areas such as hairlines, accessories, and mouth, as jittering artifacts can be identified in those areas. Additionally,  we seek to investigate the disentanglement between variations in the driving video's facial attributes such as absolute location, movement, gender, and expression and the quality of the generated video. We also plan to evaluate the effectiveness of MegActor's pipeline when integrated with a stronger video generation base model, such as SDXL.

\section{Conclusion}

We proposed MegActor, a pioneering conditional diffusion model designed for portrait animation, capable of generating facial expressions and movements consistent with the raw driving video.
Our approach introduces a synthetic data generation framework to create videos with consistent actions and expressions but inconsistent IDs, alleviating the ID leakage problem. 
Besides, we extract the foreground and background information from the reference image and inject them into the driving encoder and the text embedding module of denoising model, respectively, to eliminate the influence of irrelevant information from the driving video.
Our final model is entirely trained on public datasets, demonstrating comparable results to commercial models on generalized source portraits and driving motions, thereby validating its effectiveness.
We hope this will be helpful to the open source community's research.


\bibliographystyle{plain}
\bibliography{main}


\end{document}